\documentclass{article}

\PassOptionsToPackage{numbers, compress}{natbib}

\usepackage[final]{neurips_2025}

\usepackage[utf8]{inputenc} 
\usepackage[T1]{fontenc}    
\usepackage{hyperref}       
\usepackage{url}            
\usepackage{booktabs}       
\usepackage{amsfonts}       
\usepackage{nicefrac}       
\usepackage{microtype}      
\usepackage[dvipsnames]{xcolor}         

\usepackage{multicol}
\usepackage{algorithmic}
\usepackage{algorithm}
\usepackage{amsmath}
\usepackage{amssymb}
\usepackage{amsthm}
\usepackage{graphics}
\usepackage{epsfig}
\usepackage{subfig}
\usepackage{multirow}
\usepackage{wrapfig}
\usepackage{mathtools}
\usepackage{makecell}
\usepackage{bbm}
\usepackage{booktabs}
\usepackage{graphicx}
\usepackage{kantlipsum}
\usepackage{centernot}
\usepackage{xfrac}
\usepackage{afterpage}
\usepackage[dvipsnames,table]{xcolor}
\usepackage{tablefootnote}
\usepackage{enumitem}
\usepackage{bbding}
\definecolor{gred}{rgb}{0.859,0.267,0.216}
\definecolor{ggreen}{rgb}{0.059,0.616,0.345}

\newtheorem{definition}{Definition}

\newtheorem{proposition}{Proposition}

\definecolor{mydarkblue}{rgb}{0,0.08,0.45}
\usepackage{hyperref}
\hypersetup{
    colorlinks=true,
    linkcolor=NavyBlue,
    filecolor=magenta,      
    urlcolor=NavyBlue,
    citecolor=NavyBlue,
    }

\newcommand{\SE}{\mathrm{SE}}
\newcommand{\SO}{\mathrm{SO}}

\newcommand{\reg}{\mathrm{reg}}

\newcommand{\ac}{a}
\newcommand{\ob}{o}
\newcommand{\am}{A}
\newcommand{\w}{w}

\usepackage{xparse}
\ExplSyntaxOn
\NewDocumentCommand{\definealphabet}{mmmm}
 {
  \int_step_inline:nnn { `#3 } { `#4 }
   {
    \cs_new_protected:cpx { #1 \char_generate:nn { ##1 }{ 11 } }
     {
      \exp_not:N #2 { \char_generate:nn { ##1 } { 11 } }
     }
   }
 }
\ExplSyntaxOff
\definealphabet{bb}{\mathbb}{A}{Z}
\definealphabet{bf}{\mathbf}{A}{Z}
\definealphabet{mc}{\mathcal}{A}{Z}
\definealphabet{mf}{\mathfrak}{A}{Z}
\definealphabet{mf}{\mathfrak}{a}{z}

\newcommand\blfootnote[1]{%
  \begingroup
  \renewcommand\thefootnote{}\footnote{#1}%
  \addtocounter{footnote}{-1}%
  \endgroup
}

\title{A Practical Guide for Incorporating Symmetry in Diffusion Policy}

\author{%
Dian Wang$^1$ \quad Boce Hu$^2$ \quad Shuran Song$^1$ \quad Robin Walters$^{2\dagger}$ \quad Robert Platt$^{2\dagger}$\\
$^1$Stanford University \quad $^2$Northeastern University \\
\texttt{\url{https://sym-in-dp.github.io}}
}

\begin{document}

\maketitle

\begin{abstract}
Recently, equivariant neural networks for policy learning have shown promising improvements in sample efficiency and generalization, however, their wide adoption faces substantial barriers due to implementation complexity. Equivariant architectures typically require specialized mathematical formulations and custom network design, posing significant challenges when integrating with modern policy frameworks like diffusion-based models. In this paper, we explore a number of straightforward and practical approaches to incorporate symmetry benefits into diffusion policies without the overhead of full equivariant designs. Specifically, we investigate (i) invariant representations via relative trajectory actions and eye-in-hand perception, (ii) integrating equivariant vision encoders, and (iii) symmetric feature extraction with pretrained encoders using Frame Averaging. We first prove that combining eye-in-hand perception with relative or delta action parameterization yields inherent SE(3)-invariance, thus improving policy generalization. We then perform a systematic experimental study on those design choices for integrating symmetry in diffusion policies, and conclude that an invariant representation with equivariant feature extraction significantly improves the policy performance. Our method achieves performance on par with or exceeding fully equivariant architectures while greatly simplifying implementation. 
\end{abstract}

\section{Introduction}

Although\blfootnote{$^{\dagger}$Equal Advising} recent advancements in incorporating symmetry in robotic policy learning have shown promising results~\cite{neural_descriptor,wang2024equivariant}, the practical impact of this approach remains limited due to the significant implementation challenges. While equivariant models~\cite{e2cnn,vector_neuron} can substantially improve sample efficiency and generalization, integrating these symmetric properties into modern policy learning frameworks presents several obstacles. First, symmetry reasoning must be tailored specifically for each policy formulation, requiring different mathematical analyses and architectures across different policy frameworks like Q-learning~\cite{corl}, actor-critic methods~\cite{iclr22}, and diffusion models~\cite{brehmer2023edgi}. This creates a steep learning curve for practitioners seeking to leverage symmetry benefits. Second, state-of-the-art equivariant architectures often introduce considerable complexity with specialized layers~\cite{escnn,e3nn} and require structured inputs~\cite{vector_neuron,liao2023equiformer}, which do not naturally align well with modern policy components like diffusion-based action generation~\cite{diffpo}, eye-in-hand perception~\cite{chi2024universal}, or pretrained vision encoders. As a result, robotics researchers frequently face a difficult choice: adopting complex equivariant architectures with specialized implementation, or maintaining practical implementation concerns at the cost of symmetry benefits. This dilemma is particularly pronounced in diffusion-based visuomotor policies~\cite{diffpo}, which have emerged as a powerful paradigm for generating smooth, multimodal robot actions. Although prior works have tried to implement equivariant diffusion models~\cite{ryu2023diffusion,yang2024equibot,wang2024equivariant}, the diffusion-denoising nature significantly enhances the difficulty of incorporating equivariant structure.

\begin{wrapfigure}[20]{r}{0.52\textwidth}
\centering
\vspace{-0.4cm}
\includegraphics[width=\linewidth]{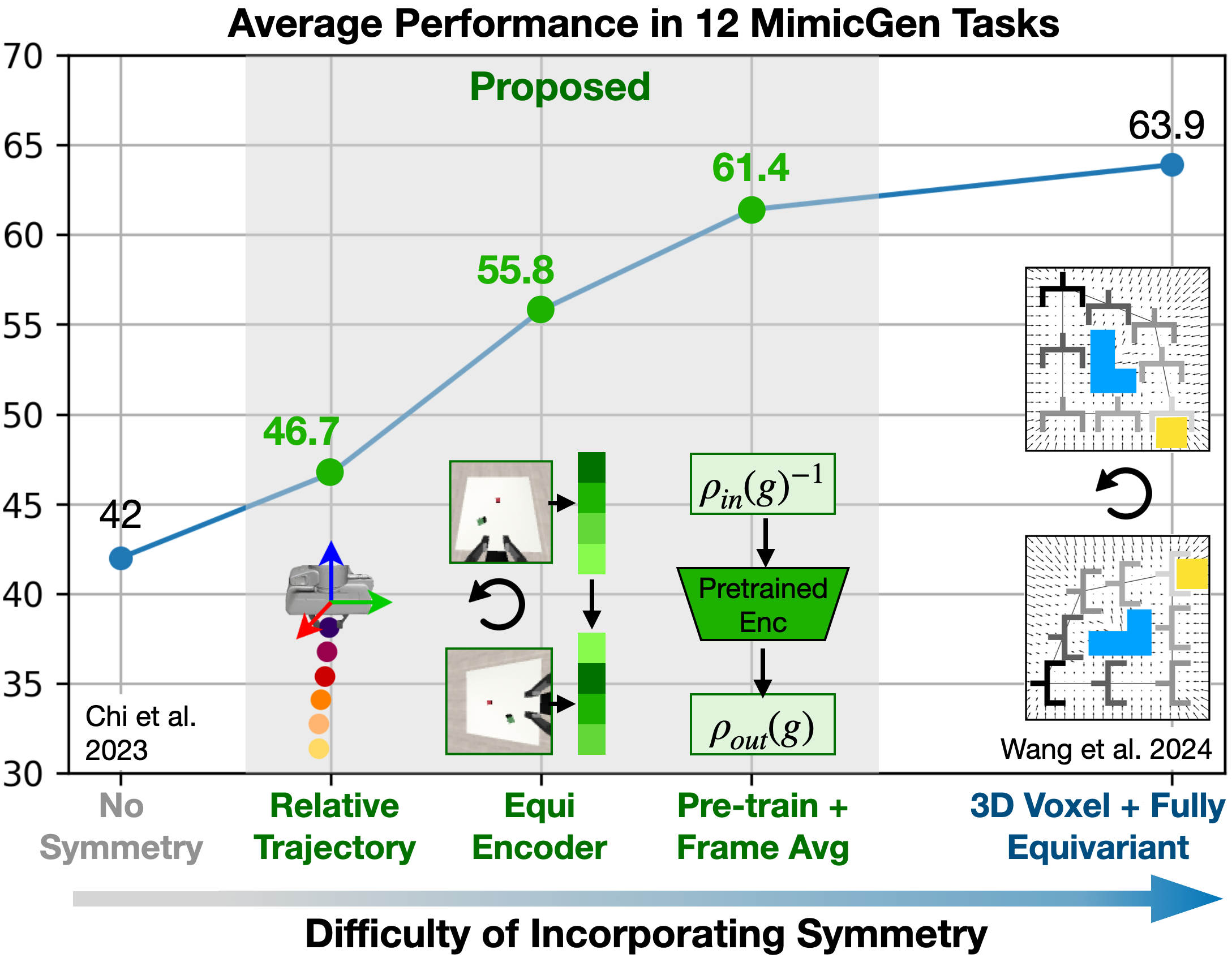}
\caption{
We propose a number of practical approaches for incorporating symmetry in Diffusion Policy, achieving comparable performance as fully-equivariant policies while maintaining simplicity.
}
\label{fig:line_chart}
\end{wrapfigure}

In this paper, we present a practical guide for incorporating symmetry into diffusion policies without requiring significant design overhead or sacrificing the advantages of modern policy formulations. We systematically investigate several straightforward approaches that achieve this balance: (i) invariant representations through relative trajectory actions and eye-in-hand perception, (ii) equivariant vision encoders that can be incorporated into standard diffusion frameworks, and (iii) Frame Averaging~\cite{puny2022frame} techniques that enable symmetrization with pretrained encoders. Our approach offers a compelling alternative to end-to-end equivariant architectures: rather than choosing between symmetry benefits and implementation practicality, our methods demonstrate that these advantages can be achieved with minimal architectural changes and overhead. As shown in Figure~\ref{fig:line_chart}, our method achieves excellent performance while maintaining implementation simplicity, positioning it at an optimal balance point in the symmetry-practicality tradeoff. Notably, our work using a single eye-in-hand image input reaches a similar performance as~\citet{wang2024equivariant}, which uses four cameras to reconstruct the 3D voxel grid input.

Our contributions can be summarized as follows:
\begin{itemize}[leftmargin=6mm]
\item We prove that eye-in-hand perception and relative trajectory action inherently possesses $\SE(3)$-invariance, significantly improving the policy's generalization.
\item We demonstrate that transitioning from absolute action representations to relative trajectory actions provides a straightforward improvement in policy performance, both with eye-in-hand perception and extrinsic perception.
\item We propose a novel approach that integrates a symmetric encoder into standard diffusion policy learning, achieved by either using equivariant network in end-to-end training or using Frame Averaging with pretrained encoders, and show that it can significantly improve performance.
\item We show that combining the invariant perception and action representations and a pretrained encoder with Frame Averaging achieves the state-of-the-art results in the MimicGen~\cite{mimicgen} benchmark while maintaining low architectural complexity and computational overhead compared to fully equivariant methods.
\end{itemize}

\section{Related Work}
\paragraph{Diffusion Policies:}
Denoising diffusion models have transformed generative modeling in vision, achieving state‐of‐the‐art results in image and video synthesis~\citep{ddpm, song2019generative} as well as planning~\citep{janner2022planning, liang2023adaptdiffuser}. Recently, \citet{diffpo,chi2023diffusion} introduced \emph{Diffusion Policy}, which extends diffusion models to robotic visuomotor control by denoising action trajectories conditioned on observations. Subsequent extensions include applications to reinforcement learning~\citep{wang2022diffusion, ren2025diffusion}, incorporation of 3D inputs~\citep{dp3, ke2024d}, hierarchical policies~\citep{xian2023chaineddiffuser, ma2024hierarchical, zhao2025hierarchical}, and large vision‐language action models~\citep{octo_2023, black2024pi_0}. A key limitation of diffusion methods is their heavy demand for training data. To mitigate this, recent works have injected domain symmetries as equivariant constraints into the denoising network, thereby boosting sample efficiency and generalization~\citep{brehmer2023edgi, ryu2023diffusion, wang2024equivariant, yang2024equibot, tie2024seed}. However, they typically require complex equivariant denoising models. In contrast, our approach integrates symmetry by combining invariant observation and action representations with an equivariant vision encoder, greatly simplifying implementation.

\paragraph{Equivariant Policy Learning:}
Robotic policies often require generalizing across spatial transformations of the environment. Traditional methods often achieve this via extensive data augmentation~\cite{zeng2018learning,transporter}. Recently, \emph{equivariant models}, a class of methods that are mathematically constrained to be equivariant~\cite{g_conv,steerable_cnns,e2cnn,escnn,vector_neuron,e3nn,liao2023equiformer,liao2024equiformerv}, has been widely adapted to robotics to automatically instantiate spatial generalization. Such models have been widely applied across robot learning, including equivariant reinforcement learning~\cite{corl,iclr22,corl22,nguyen2023equivariant,nguyen2024symmetry,kohler2023symmetric,liu2023continual,hoang2025geometryaware}, imitation learning~\cite{jia2023seil,yang2024equivact,gao2024riemann}, grasp learning~\cite{rss22xupeng,zhu2023robot,huang2023edge,huorbitgrasp,lim2024equigraspflow}, and pick-place policies~\cite{neural_descriptor,simeonov2023se,pan2023tax,ryu2023equivariant,rss22haojie,huang2024fourier,huang2023leveraging,huang2024imagination,eisner2024deep,huang2024match}. However, these methods often demand complex symmetry reasoning and equivariant layers, which can hinder scalability. By contrast, our framework introduces symmetry in a more modular fashion, making it easier to implement and adapt.

\paragraph{Policy Learning using Eye-in-Hand Images:}
Eye‐in‐hand perception, using a camera mounted on the robot’s end‐effector, has been a popular choice in manipulation because it is simple and calibration‐free. For instance, \citet{jangir2022look} learn a shared latent space between egocentric and external views to train hybrid‐input policies. \citet{hsu2022visionbased} show that eye‐in‐hand images (alone or combined with external cameras) yield higher success rates and better generalization, and similar findings have been captured in~\citep{robomimic}. Consequently, many recent frameworks retain eye‐in‐hand imagery~\citep{aloha, aldaco2024aloha, zhao2024aloha, black2024pi_0, liu2025rdtb}. Another advantage is rapid data collection using a handheld gripper~\citep{song2020grasping, young2021visual, pari2021surprising, chi2024universal, wu2024fast, ha2024umi}. In this work, we theoretically analyze how eye‐in‐hand observation, when paired with relative or delta trajectory actions, yields inherent symmetry advantages for diffusion‐based policies.


\section{Background}
\textbf{Problem Statement:}
We consider behavior cloning for visuomotor policy learning in robotic manipulation, where the goal is to learn a policy that maps an observation $\ob$ to an action $\ac$, mimicking an expert policy. Both $\ob$ and $\ac$ may span multiple time steps, i.e., $\ob = \{\ob_{t-(m-1)}, \dots, \ob_{t-1}, \ob_t\}, \ac=\{\ac_t, \ac_{t+1}, \dots, \ac_{t+(n-1)} \}$, where $m$ is the number of past observations and $n$ is the number of future action steps. 
At time step $t$, the observation $\ob_t=(I_t, T_t, w_t)$ contains the visual information $I_t$, (e.g., images), the pose of the gripper in the world frame $T_t\in\SE(3)$, as well as the gripper aperture $\w_t\in \bbR$. The action $\ac_t=(\am_t, \w_t)$ specifies a target pose $\am_{t}\in\SE(3)$ of the gripper and an open-width command $\w_t\in \bbR$. To simplify the notation for our analysis, we omit the gripper command $w_t$ in the action and focus on the pose command by writing $a=\{A_t, A_{t+1} \dots,A_{t+n-1}\}$ (while in the actual implementation the policy controls both the pose and the aperture).


\textbf{Diffusion Policy:}
\citet{diffpo} introduced Diffusion Policy, which formulates the behavior cloning problem as learning a Denoising Diffusion Probabilistic Model (DDPMs)~\citep{ddpm} over action trajectories.
Diffusion Policy learns a noise prediction function $\varepsilon_\theta(\ob, \ac+\varepsilon^k, k)=\varepsilon^k$ using a network $\varepsilon_\theta$, which is trained to predict the noise \(\varepsilon^k\) added to an action $\ac$. 
The training loss is $\mathcal{L}=||\varepsilon_\theta(\ob, \ac + \varepsilon^k, k) - \varepsilon^k||^2$, where $\varepsilon^k$ is a random noise conditioned on a randomly sampled denoising step $k$.
At inference, starting from $\ac^k\sim \mathcal{N}(0, 1)$, the model iteratively denoises
\begin{equation}
\label{eqn:ddpm}
    \ac^{k-1}=\alpha(\ac^k - \gamma \varepsilon_\theta(\ob, \ac^k, k) + \epsilon),
\end{equation}
where $\epsilon\sim \mathcal{N}(0, \sigma^2I)$. $\alpha, \gamma, \sigma$ are functions of the denoising step $k$ (also known as the noise schedule). The action $\ac^0$ is the final executed action trajectory.

\textbf{Equivariance:}
\label{sec:bk_equi}
A function \(f\colon X \to Y\) is \emph{equivariant} to a group \(G\) if, for all \(g\in G\),
\[
f\bigl(\rho_x(g)\,x\bigr)
= \rho_y(g)\,f(x),
\]
where \(\rho_x\) and \(\rho_y\) are representations of \(G\) on \(X\) and \(Y\).  Equivariance ensures that applying \(g\) before or after \(f\) yields the same result, thus $f$ is $G$-symmetric. When $\rho$ is clear, we often write $g \cdot x$ or $gx$.

For 2D images, the group $\SO(2)=\{\mathrm{Rot}_\theta:0\leq \theta < 2\pi\}$ of planar rotations and its subgroup $C_u=\{\mathrm{Rot}_\theta: \theta \in \{\frac{2\pi i}{u}|0\leq i <u\}\}$ of rotations by multiples of $\frac{2\pi}{u}$ are often used to construct equivariant neural networks~\cite{e2cnn,escnn} that can capture rotated features. The \emph{regular representation} $\rho_\reg:G\to \mathbb{R}^{u\times u}$ of $C_u$ is of particular interest of this paper, which defines how $C_u$ acts on a vector $x\in \mathbb{R}^{u}$ by $u\times u$ permutation matrices.  Intuitively, the vector $x$ can be viewed as containing information for each rotation in $C_u$. Let $r^v \in C_u=\{1, r^1, \dots r^{u-1}\}$ and $x=(x_1, \dots x_u) \in \mathbb{R}^u$, then $\rho_\reg(r^v)x=(x_{u-v+1}, \dots,x_u, x_1, x_2, \dots, x_{u-m})$ cyclically permutes the coordinates of $x$.

\section{Approaches for Incorporating Symmetry in Diffusion Policy}

\begin{figure}[t]
\centering
\includegraphics[width=\linewidth]{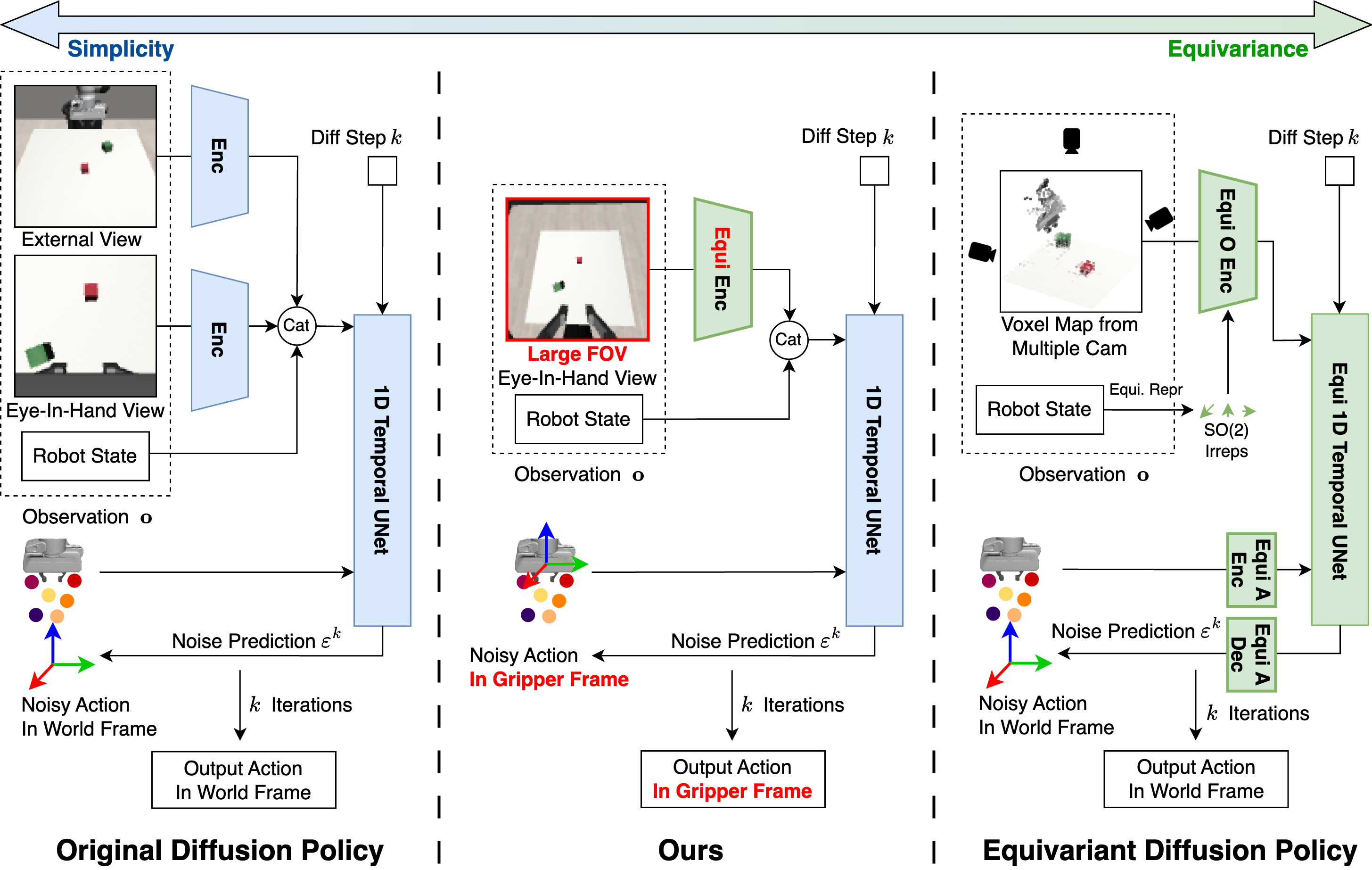}
\caption{The difference comparing our method (middle) with the original Diffusion Policy (left) and the Equivariant Diffusion Policy (right). We use only an eye-in-hand image as the input, and use an equivariant encoder to acquire symmetry-aware features from the input image. In policy denoising, the noisy action and the noise-free action output are both in the gripper frame. Other components remain identical to the original Diffusion Policy. Compared with Equivariant Diffusion Policy (right), our approach is significantly simpler while maintaining a comparable experimental performance.}
\label{fig:vs_diff}
\end{figure}

In this section, we introduce three practical approaches for incorporating symmetry into diffusion policies without requiring complex end-to-end equivariant architecture design. First, we examine how invariant action and perception representations naturally induce symmetric properties. Second, we explore integrating equivariant vision encoders that extract symmetry-aware features while maintaining standard diffusion heads. Finally, we present how to leverage pre-trained vision encoders in an equivariant way through Frame Averaging~\cite{puny2022frame}. Together, these approaches offer a spectrum of options for balancing symmetry benefits with implementation simplicity. As shown in Figure~\ref{fig:vs_diff}, our proposed approach (middle) requires minimal architectural change compared with the original Diffusion Policy~\cite{diffpo} (left), and is much simpler than the fully equivariant model~\cite{wang2024equivariant} (right).

\subsection{Representing Actions as Absolute, Relative, and Delta Trajectory}

\label{sec:inv_repr}
Choosing the right representation for actions and observations is crucial for sample-efficient policy learning in robotic manipulation. When a robotic task and environment exhibit rigid-body symmetries, incorporating an equivariant or invariant representation can significantly enhance generalization to unseen object configurations and poses. In this section, we explore three natural trajectory action representations: absolute, relative, and delta trajectories, highlighting their symmetry properties under global $\SE(3)$ transformations. Notice that although relative trajectories were introduced by \citet{chi2024universal}, their symmetric advantages have not yet been explored.

\begin{definition}[Absolute Trajectory Action]
An absolute trajectory action specifies future gripper poses directly in the world frame as
\[
a = \bigl\{A_{t},A_{t+1},\dots,A_{t+n-1}\bigr\},
\]
where each $A_{t+i}\in \SE(3)$ is the desired pose at time $t+i$.  
\end{definition}


\begin{definition}[Relative Trajectory Action]
\label{def:rel-traj}
Let $T_t\in \SE(3)$ be the current gripper pose in the world frame.  A \emph{relative trajectory action} is a sequence
\[
a^r = \bigl\{A_{t}^{r},A_{t+1}^{r},\dots,A_{t+n-1}^{r}\bigr\},
\]
where each $A_{t+i}^{r}\in \SE(3)$ specifies the gripper’s pose relative to its initial frame at time $t$.  The corresponding absolute poses are recovered via
\begin{equation}
\label{eqn:rel_to_abs}
A_{t+i} = T_t\,A_{t+i}^{r},\qquad i=0,\dots,n-1.
\end{equation}
\end{definition}

\begin{definition}[Delta Trajectory Action]
A \emph{delta trajectory action} is a sequence of incremental transforms expressed in a moving local frame:
\[
a^d = \bigl\{A_{t}^{d},A_{t+1}^{d},\dots,A_{t+n-1}^{d}\bigr\},
\]
where $A_{t+i}^{d}\in SE(3)$ represents the incremental motion at time step $t+i$ expressed relative to the gripper’s frame at the previous time step $t+i-1$. The absolute poses are reconstructed as:
\begin{equation}
\label{eqn:delta_to_abs}
A_{t+i}
=\;T_t\;\Bigl(\prod_{j=0}^{i-1}A_{t+j}^{d}\Bigr),\qquad i=0,\dots,n-1.
\end{equation}
\end{definition}

We now formalize the transformation properties of these action representations (see Appendix~\ref{app:proof_action} for the proof):
\begin{proposition}[Equivariance and Invariance under $\SE(3)$]
\label{prop:action}
Consider a global transformation $g \in \SE(3)$ applied to the world coordinate frame, which transforms the current gripper pose as $T_t \mapsto g T_t$. Under this transformation:

\begin{enumerate}[leftmargin=6mm]
\item The absolute trajectory action transforms {equivariantly}, i.e., $g\cdot a \;=\;\{gA_{t},\,gA_{t+1},\dots\}\,$.

\item The relative trajectory action is {invariant}, i.e., $g\cdot a^r = a^r$.

\item The delta trajectory action is {invariant}, i.e., $g\cdot a^d = a^d$.
\end{enumerate}
\end{proposition}

\subsection{SE(3)-Invariant Policy Learning}
\label{sec:inv_policy}

\begin{wrapfigure}[21]{r}{0.5\textwidth}
\centering
\vspace{-0.3cm}
\includegraphics[width=\linewidth]{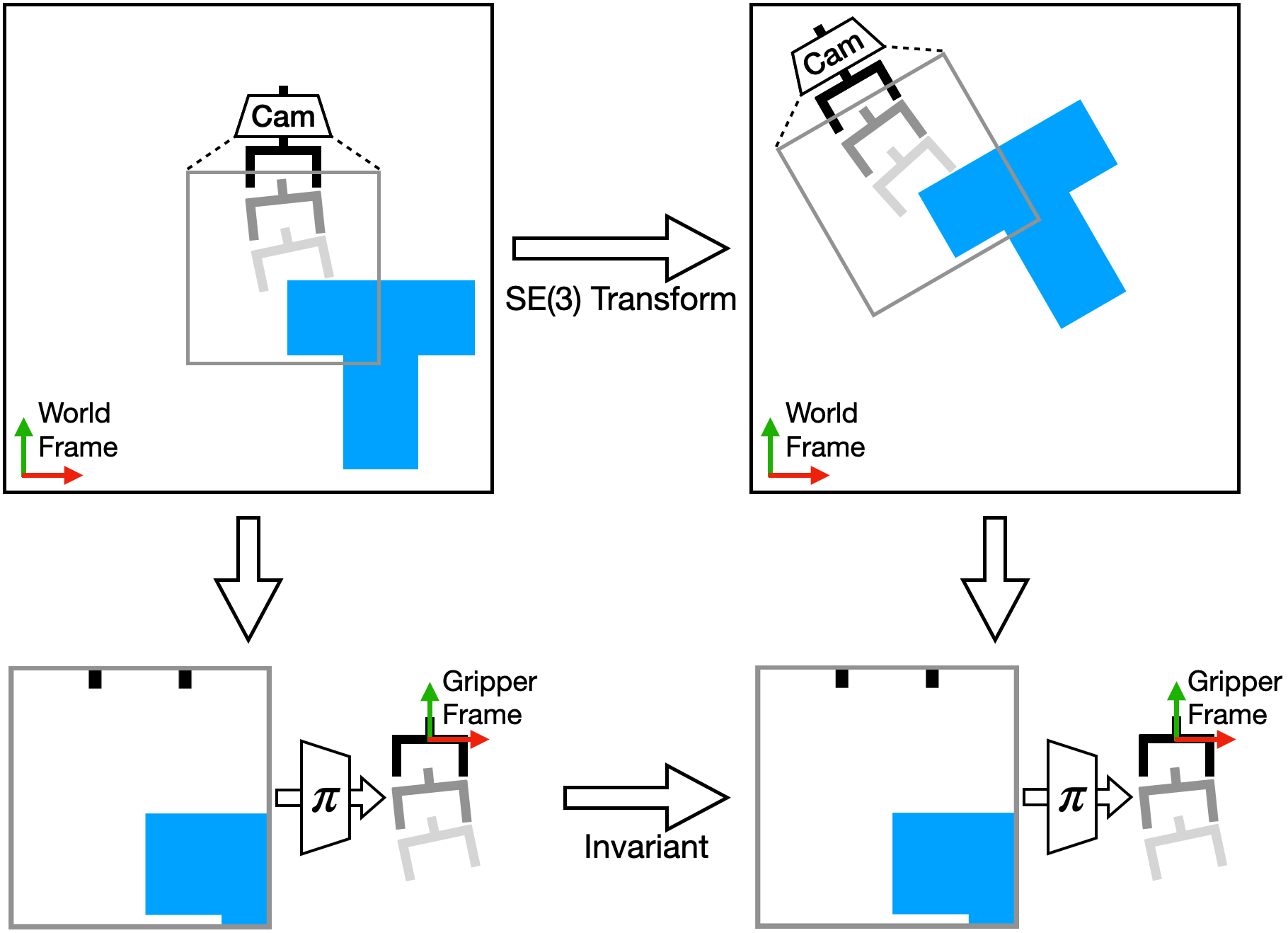}
\caption{The invariant property of an eye-in-hand perception and action representation in policy learning. Top: an $\SE(3)$ transform applied to the world, the action should transform accordingly. Bottom: for the policy, both the input eye-in-hand image and the output relative trajectory (in the gripper frame) remain invariant.}
\label{fig:sym}
\end{wrapfigure}

A key advantage of using relative or delta trajectory action representations is that, if the policy being modeled is equivariant, i.e., $\pi(go) = g\pi(o)$, the learned function $\bar{\pi}:\ob \mapsto \ac^r$ or $\bar{\pi}:\ob \mapsto \ac^d$ becomes invariant, i.e., $\bar\pi(go) = \bar{\pi}(o)$. This makes the underlying denoising network $\epsilon_\theta$ also invariant, 
significantly reducing the function space and potentially easing the training.

Moreover, when the relative or delta trajectory is combined with eye-in-hand perception, it naturally yields an $\SE(3)$-invariant canonicalization. 
Specifically, consider an agent equipped with a gripper-mounted camera, which captures an eye-in-hand image $I_t$. When an arbitrary transformation $g \in \SE(3)$ is applied to the world, the visual input from the eye-in-hand camera remains unchanged since the relative position and orientation between the camera and the world do not vary. For the input observation $\ob_t = (I_t, T_t, w_t)$ consisting of the invariant image $I_t$, a gripper pose $T_t$, and gripper open-width $w_t$, applying the transformation $g$ to the world frame affects only the gripper pose $T_t$,
\begin{equation}
\label{eqn:go}
g \cdot \ob_t = (I_t, gT_t, w_t).
\end{equation}

Let us first assume that the policy does not depend on the gripper pose $T_t$, then the policy $\pi\colon \ob \mapsto \ac$ is $\SE(3)$-equivariant when using eye-in-hand perception and relative or delta action:

\begin{proposition}
\label{prop:policy}
Let $\bar{\pi}: \ob \mapsto \ac^r$ or $\bar{\pi}: \ob \mapsto \ac^d$ be a function mapping from the observation $\ob$ to the relative trajectory $\ac^r$ or the delta trajectory $\ac^d$. Assume an eye-in-hand observation is used where Equation~\ref{eqn:go} is satisfied and $\bar{\pi}$ does not depend on $T_t$. If the policy $\pi:\ob \mapsto \ac$ reconstructs the absolute trajectory $\ac$ using Equation~\ref{eqn:rel_to_abs} or~\ref{eqn:delta_to_abs}, then $\pi$ is $\SE(3)$-equivariant, i.e., $\pi(g\ob) = g\pi (\ob)$.
\end{proposition}

See Appendix~\ref{app:proof_policy} for the proof. This equivariance property implies that once a policy is learned, it automatically generalizes across different poses in space without additional training data, thus enhancing sample efficiency and robustness. As shown in Figure~\ref{fig:sym}, when an $\SE(3)$ transformation is applied to the world, both the perception and action remain invariant. 

In practice, the assumption that $\bar{\pi}$ does not depend on $T_t$ will not perfectly hold because the changing $T_t$ will affect the network prediction, thus we will only have an approximate invariance property. Since $T_t$ provides important information to the policy despite breaking the symmetry, we choose not to explicitly constrain the policy to be invariant to $T_t$. Still, our experiments demonstrate significant performance improvements when employing this approximate invariant property.

\subsection{Equivariant Vision Encoders}

While the invariant representations described in Section~\ref{sec:inv_policy} theoretically achieve $\SE(3)$-equivariant policy learning, we experimentally found that they do not fully match the performance of end-to-end equivariant policies~\cite{wang2024equivariant}. This is because equivariant neural networks not only guarantee global symmetric transformations, but more importantly, they extract richer local features that can capture the underlying symmetries of the problem domain. Instead of falling back to a network architecture that is end-to-end equivariant, we propose a novel approach that incorporates an equivariant vision encoder to extract symmetry-aware features while preserving a standard non-equivariant diffusion backbone. This approach would preserve the benefits of symmetric feature extraction without the complexity of a full equivariant model. 

Specifically, we can replace the standard CNN vision encoder in a Diffusion Policy with an equivariant CNN that operates on the group $C_u \subset \SO(2)$. This encoder maps the input eye-in-hand image to a feature vector that transforms according to the regular representation of $C_u$, providing a richer representation to the diffusion head with explicit information about how features transform under rotations, significantly enhancing learning.

\subsubsection{Incorporating Pretrained Vision Encoders with Frame Averaging} 

Although there exists a wide variety of equivariant neural network architectures~\cite{g_conv,e2cnn,escnn}, they typically require defining each layer to be equivariant by constraining the weights with specialized kernels~\cite{equi_theory}. This usually implies that the network is built specifically for a task and is trained from scratch. However, modern computer vision has seen tremendous progress through large-scale pretrained models, which provide powerful general-purpose representations. 
To bridge the gap between equivariance and pre-training, we employ \emph{Frame Averaging}~\cite{puny2022frame}  for turning an arbitrary function $\Phi:X\to Y$ into a $G$-equivariant network by averaging over a \emph{Frame} $\mathcal{F}:X\to 2^G \setminus\{\emptyset\} $ that satisfies equivariance as a set $\mathcal{F}(gx) = \mathcal{F}(x)$:
\begin{equation}
\label{eqn:frame_avg}
\Psi(x)=\frac{1}{|\mathcal{F}(x)|}\sum_{g\in \mathcal{F}(x)}\rho_y(g)\Phi(\rho_x(g)^{-1}x),
\end{equation}
where $\Psi: X\to Y$ will have the equivariant property $\Psi(\rho_x(g)x)=\rho_y(g)\Psi(x)$. For a finite group $G$, one can set the frame to be the whole group $\mathcal{F}(x)=G$, and Equation~\ref{eqn:frame_avg} becomes \emph{symmetrization}:
\begin{equation}
\label{eqn:symmetrization}
\Psi(x)=\frac{1}{|G|}\sum_{g\in G}\rho_y(g)\Phi(\rho_x(g)^{-1}x).
\end{equation}
When using a pretrained encoder with Frame Averaging, we obtain the benefits of both powerful pretrained representations and explicit rotational equivariance, allowing us to leverage state-of-the-art vision backbones without sacrificing symmetry properties.


\section{Experiments}

\begin{figure*}[t]
  \newlength{\env}
  \setlength{\env}{0.11\linewidth}
  \centering
  \subfloat[Stack D1]{
  \includegraphics[width=\env]{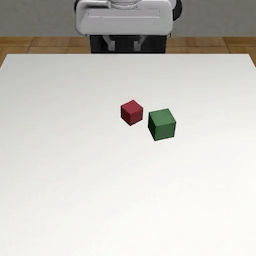}
  \includegraphics[width=\env]{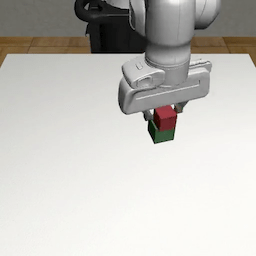}
  }
  \subfloat[Stack Three D1]{
  \includegraphics[width=\env]{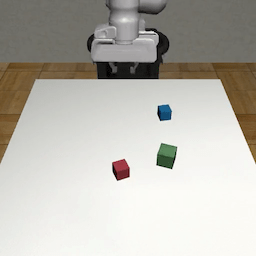}
  \includegraphics[width=\env]{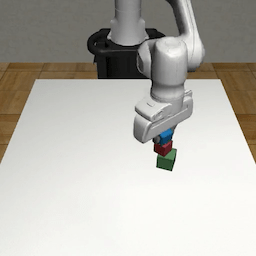}
  }
  \subfloat[Square D2]{
  \includegraphics[width=\env]{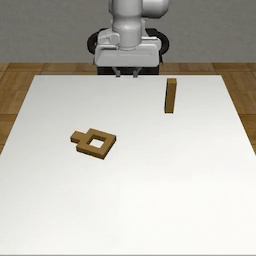}
  \includegraphics[width=\env]{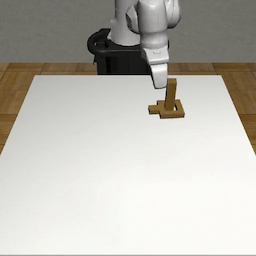}
  }
  \subfloat[Threading D2]{
  \includegraphics[width=\env]{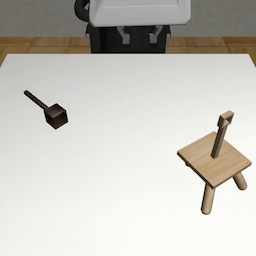}
  \includegraphics[width=\env]{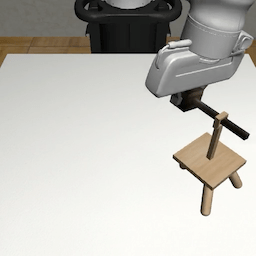}
  }\\
  \subfloat[Coffee D2]{
  \includegraphics[width=\env]{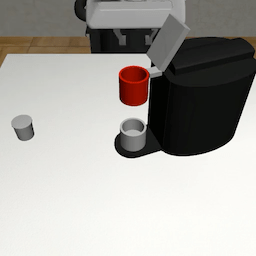}
  \includegraphics[width=\env]{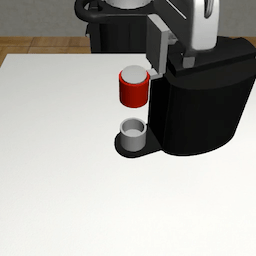}
  }
  \subfloat[Three Pc. Assembly D2]{
  \includegraphics[width=\env]{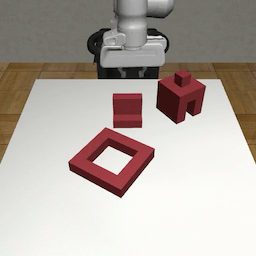}
  \includegraphics[width=\env]{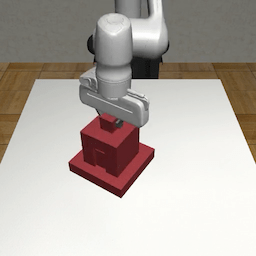}
  }
  \subfloat[Hammer Cleanup D1]{
  \includegraphics[width=\env]{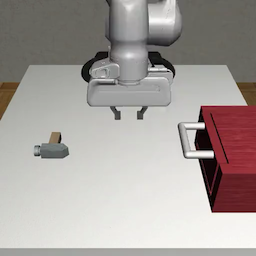}
  \includegraphics[width=\env]{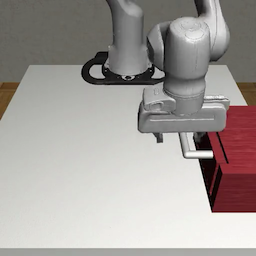}
  }
  \subfloat[Mug Cleanup D1]{
  \includegraphics[width=\env]{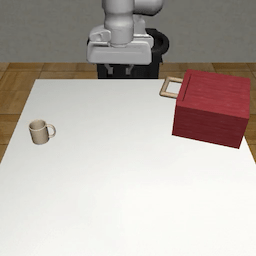}
  \includegraphics[width=\env]{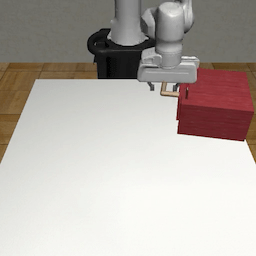}
  }\\
  \subfloat[Kitchen D1]{
  \includegraphics[width=\env]{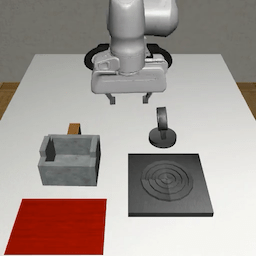}
  \includegraphics[width=\env]{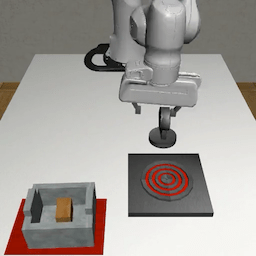}
  }
  \subfloat[Nut Assembly D0]{
  \includegraphics[width=\env]{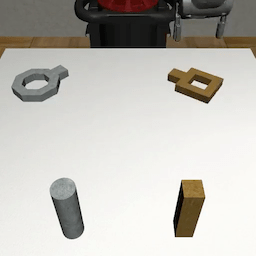}
  \includegraphics[width=\env]{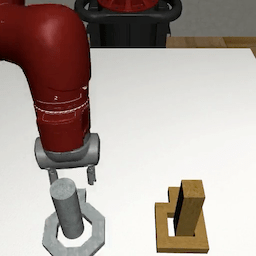}
  }
  \subfloat[Pick Place D0]{
  \includegraphics[width=\env]{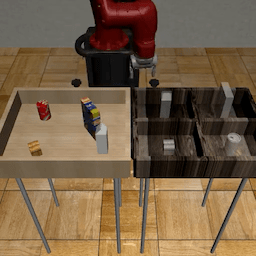}
  \includegraphics[width=\env]{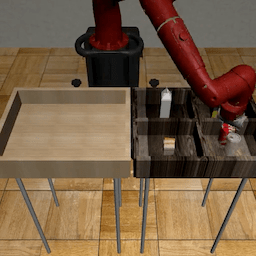}
  }
  \subfloat[Coffee Preparation D1]{
  \includegraphics[width=\env]{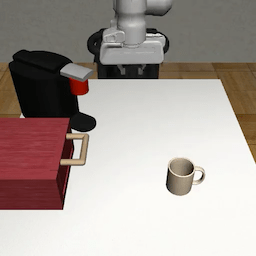}
  \includegraphics[width=\env]{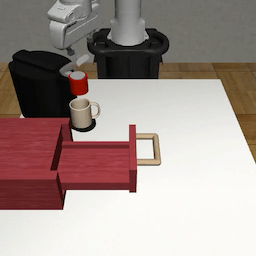}
  }
  \caption{The experimental environments from MimicGen~\citep{mimicgen}. The left image in each subfigure shows the initial state of the environment; the right image shows the goal state.
  }
  \label{fig:envs}
  \end{figure*}

In this section, we conduct a systematic experimental study comparing different approaches for incorporating symmetry in diffusion policies. We investigate the following key research questions:
\begin{enumerate}[leftmargin=6mm]
\item \textbf{Invariant representations:} How do the invariant action and perception representations analyzed in Section~\ref{sec:inv_repr} impact diffusion policy learning performance?
\item \textbf{Equivariant vision encoders:} Can diffusion policies benefit from incorporating equivariant vision encoders?
\item \textbf{Pre-trained encoders:} How effectively can we leverage pre-trained encoders with Frame Averaging (Equation~\ref{eqn:frame_avg})?
\item \textbf{Comparison to end-to-end equivariant diffusion:} How do these approaches compare with fully equivariant diffusion policies~\cite{wang2024equivariant}?
\end{enumerate}

We evaluate our approaches on 12 robotic manipulation tasks in the MimicGen~\citep{mimicgen} benchmark, as illustrated in Figure~\ref{fig:envs}. We perform an additional Robomimic~\cite{robomimic} experiment in Appendix~\ref{app:robomimic}.

\subsection{Action and Observation Representation}

We first evaluate the three action representations (absolute trajectory, relative trajectory, and delta trajectory) discussed in Section~\ref{sec:inv_repr} across two different observation settings, \emph{Large FOV In-Hand} and \emph{In-hand + External}. Figure~\ref{fig:vs_diff} illustrates the differences between these configurations. 

\begin{wrapfigure}[12]{r}{0.35\textwidth}
\centering
\vspace{-0.6cm}
\subfloat[Threading]{
\includegraphics[width=0.46\linewidth]{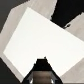}
}
\subfloat[Coffee Prep.]{
\includegraphics[width=0.46\linewidth]{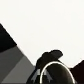}
}
\caption{The failures of the in-hand observation due to occlusion and insufficient information.}
\label{fig:failure_case}
\end{wrapfigure}

The results, presented in Table~\ref{tab:act_obs_repr}, demonstrate several key findings. First, relative trajectory consistently outperforms absolute trajectory in 10 out of 12 tasks in both Large FOV In-Hand and In-Hand + External observations. On average, relative trajectory provides a 5.9\% improvement over absolute trajectory with Large FOV In-Hand observations and a 7.4\% improvement with In-Hand + External observations. These results align with our theoretical analysis in Section~\ref{sec:inv_repr}, confirming that the symmetry properties of relative trajectory representations contribute to better performance. However, despite having the same theoretical guarantee, delta trajectory empirically performs poorly, underperforming absolute trajectory by 2.9\% on average, where it only performs well in relatively simple tasks. We hypothesize that this is because delta trajectory can be interpreted as a sequence of velocity vectors, containing less temporal and structural information for the denoising process. Notice that similar observations of the underperformance of velocity control in diffusion policy learning were also reported in prior works~\cite{diffpo,wang2024equivariant}.

When comparing across observation settings using relative trajectory, we find that Large FOV In-Hand generally performs better or on par with In-Hand + External. However, if averaged across all tasks, the Large FOV In-Hand setup underperforms by 2.7\%. This performance gap is primarily due to a significant drop in the Coffee Preparation task, where the eye-in-hand view alone provides insufficient information for completing this long-horizon task. Moreover, we found that tasks like Threading sometimes encounter occlusion challenges. As shown in Figure~\ref{fig:failure_case}, those limitations of a single eye-in-hand image constitute the majority of the failure modes. Despite these drawbacks, leveraging the invariant observation and action representations provides a 4.7\% improvement compared with the original Diffusion Policy, which uses In-Hand + External views and absolute trajectory.

\begin{table*}[t]
\scriptsize
\setlength\tabcolsep{4pt}
\centering
\setlength{\fboxsep}{1pt}
\caption{The performance comparison between different action representations and different observation setups. All policies are trained using 100 expert demonstrations. We perform 60 evaluations (each with 50 policy rollouts) throughout the training, and report the best average success rate. Results averaged over three seeds, $\pm$ indicates standard error. \colorbox{cyan!10}{Blue box} indicates relative trajectory or delta trajectory outperforming absolute trajectory in the corresponding observation setting; \colorbox{red!10}{Red box} indicates underperforming. \textbf{Bold} indicates best performing method across different settings. Performance of Abs Traj in In-Hand + External is reported by~\cite{wang2024equivariant}.
}
\label{tab:act_obs_repr}
\begin{tabular}{l l c c c c c c c}
\toprule
Obs & Action & Mean & Stack D1 & Stack Three D1 & Square D2 & Threading D2 & Coffee D2 & Three Pc. D2 \\
\midrule
\multirow{3}{*}{\begin{tabular}[l]{@{}l@{}}Large FOV\\In-Hand\end{tabular}} & Rel Traj & 46.7 & \cellcolor{cyan!10}\textbf{98.0$\pm$0.0} & \cellcolor{cyan!10}\textbf{72.0$\pm$1.2} & \cellcolor{cyan!10}16.0$\pm$1.2 & \cellcolor{cyan!10}16.0$\pm$1.2 & \cellcolor{cyan!10}56.7$\pm$1.8 & \cellcolor{cyan!10}\textbf{4.0$\pm$1.2} \\
 & Delta Traj & 37.9 & \cellcolor{cyan!10}96.7$\pm$0.7 & \cellcolor{cyan!10}54.7$\pm$3.5 & \cellcolor{cyan!10}10.0$\pm$1.2 & \cellcolor{cyan!10}11.3$\pm$1.3 & \cellcolor{red!10}38.7$\pm$0.7 & 2.0$\pm$1.2 \\
& Abs Traj & 40.8 & 94.0$\pm$1.2 & 52.7$\pm$2.4 & 7.3$\pm$0.7 & 10.0$\pm$0.0 & 44.0$\pm$4.6 & 2.0$\pm$0.0 \\
\midrule
\multirow{2}{*}{\begin{tabular}[l]{@{}l@{}}In-Hand +\\External\end{tabular}} & Rel Traj & 49.4 & \cellcolor{cyan!10}89.3$\pm$2.7 & \cellcolor{cyan!10}52.7$\pm$3.7 & \cellcolor{cyan!10}\textbf{20.7$\pm$1.3} & \cellcolor{cyan!10}\textbf{18.7$\pm$3.5} & \cellcolor{cyan!10}\textbf{63.3$\pm$1.3} & \cellcolor{red!10}3.3$\pm$0.7 \\
& Abs Traj & 42.0 & 76.0$\pm$4.0 & 38.0$\pm$0.0 & 8.0$\pm$1.2 & 17.3$\pm$1.8 & 44.0$\pm$1.2 & \textbf{4.0$\pm$0.0} \\
\midrule
\midrule
Obs & Method & & Hammer Cl. D1 & Mug Cl. D1 & Kitchen D1 & Nut Asse. D0 & Pick Place D0 & Coffee Prep. D1 \\
\midrule
\multirow{3}{*}{\begin{tabular}[l]{@{}l@{}}Large FOV\\In-Hand\end{tabular}} & Rel Traj & & \cellcolor{red!10}{60.7$\pm$0.7} & \cellcolor{cyan!10}\textbf{50.7$\pm$1.8} & \cellcolor{cyan!10}\textbf{68.7$\pm$3.7} & \cellcolor{cyan!10}44.7$\pm$3.3 & \cellcolor{cyan!10}\textbf{42.7$\pm$1.6} & 30.7$\pm$1.8 \\
& Delta Traj & & \cellcolor{cyan!10}\textbf{62.7$\pm$2.4} & \cellcolor{cyan!10}47.3$\pm$1.8 & \cellcolor{red!10}57.3$\pm$4.8 & \cellcolor{red!10}17.0$\pm$1.5 & \cellcolor{red!10}31.7$\pm$0.4 & \cellcolor{red!10}26.0$\pm$1.2 \\
& Abs Traj & & 62.0$\pm$2.0 & 44.7$\pm$1.8 & 66.7$\pm$0.7 & 43.0$\pm$2.5 & 32.0$\pm$0.6 & 30.7$\pm$2.9 \\
\midrule
\multirow{2}{*}{\begin{tabular}[l]{@{}l@{}}In-Hand +\\External\end{tabular}} & Rel Traj & & \cellcolor{cyan!10}58.7$\pm$0.7 & \cellcolor{cyan!10}46.7$\pm$0.7 & \cellcolor{red!10}62.0$\pm$2.0 & \cellcolor{cyan!10}\textbf{57.3$\pm$0.9} & \cellcolor{cyan!10}39.8$\pm$1.2 & \cellcolor{cyan!10}\textbf{80.0$\pm$2.0} \\
& Abs Traj & & 52.0$\pm$1.2 & 42.7$\pm$0.7 & 66.7$\pm$2.4 & 54.7$\pm$2.3 & 35.3$\pm$2.2 & 65.3$\pm$0.7 \\
\bottomrule
\end{tabular}
\end{table*}

\subsection{Integrating Symmetry into the Vision Encoder}
\label{sec:exp_equi_enc}

Having established the advantages of invariant action representations, we now investigate different approaches for incorporating symmetry into the vision encoder component of diffusion policies. We compare four methods: \textit{CNN Encoder (CNN Enc):} A standard ResNet-18~\cite{resnet} without any symmetry constraints, trained from scratch; \textit{Equivariant Encoder (Equi Enc):} An equivariant ResNet-18 architecture implemented with equivariant layers using the escnn~\cite{escnn} library, enforcing $C_8$-equivariance with outputs in the regular representation of $C_8$; \textit{Pretrained Encoder (Pretrain):} A standard ResNet-18 pretrained on ImageNet-1k~\cite{deng2009imagenet}, without explicit symmetry constraints; \textit{Pretrained Encoder with Frame Averaging (Pretrain + FA):} A pretrained ResNet-18 enhanced with Frame Averaging (Equation~\ref{eqn:frame_avg}) to achieve $C_8$-equivariance without modifying the underlying network architecture.

Table~\ref{tab:equi_enc} presents our findings across all 12 manipulation tasks. The results reveal several important insights:
First, comparing non-pretrained encoders (Equi Enc vs. CNN Enc), we observe that incorporating equivariance improves performance in 11 out of 12 tasks, yielding a substantial 9.1\% average improvement. This confirms that explicit symmetry constraints significantly benefit diffusion policy learning. Second, in the pretrained encoder setting, adding Frame Averaging (Pretrain + FA vs. Pretrain) leads to a 4.1\% average performance improvement, with superior results in 7 out of 12 tasks. This demonstrates that symmetry benefits can be obtained even when leveraging powerful pretrained representations. Third, comparing our approaches to Equivariant Diffusion Policy~\cite{wang2024equivariant} (EquiDiff), we find that both Equi Enc and Pretrain + FA achieve competitive performance. 

Specifically, our Equi Enc approach outperforms image-based EquiDiff (Im) on average, while Pretrain + FA achieves results only 2.5\% below voxel-based EquiDiff (Vo). This is particularly impressive considering that EquiDiff (Vo) utilizes RGBD inputs from four cameras and employs a substantially more complex architecture. In contrast, our Pretrain + FA approach requires only a single eye-in-hand RGB image and minimal equivariant reasoning, making it considerably more practical for real-world deployment.

Overall, these results suggest that integrating symmetry through equivariant encoders provides significant performance benefits for diffusion policies, with Frame Averaging offering an elegant way to leverage powerful pretrained representations while maintaining equivariance properties.

\begin{table*}[t]
\scriptsize
\setlength\tabcolsep{1.5pt}
\centering
\setlength{\fboxsep}{1pt}
\caption{The performance comparison between symmetric encoder and standard encoder. All policies are trained using 100 expert demonstrations. We perform 60 evaluations (each with 50 policy rollouts) throughout the training, and report the best average success rate. Results averaged over three seeds, $\pm$ indicates standard error. \colorbox{green!30}{Dark green box} indicates outperforming EquiDiff with voxel inputs; \colorbox{green!10}{Light green box} indicates outperforming EquiDiff with image inputs; \colorbox{yellow!20}{Yellow box} indicates underperforming both. \textbf{Bold} indicates whether the symmetric encoder outperforms the non-symmetric version in the corresponding training setting. Performance of EquiDiff is reported by~\cite{wang2024equivariant}.}
\label{tab:equi_enc}
\begin{tabular}{l l l c c c c c c c}
\toprule
Obs & Action & Method & Mean & Stack D1 & Stack Three D1 & Square D2 & Threading D2 & Coffee D2 & Three Pc. D2 \\
\midrule
\multirow{4}{*}{\begin{tabular}[l]{@{}l@{}}Large FOV\\In-Hand\end{tabular}} & \multirow{4}{*}{Rel Traj} & Pretrain + FA & {61.4} & \cellcolor{green!30}\textbf{100.0$\pm$0.0} & \cellcolor{green!30}\textbf{86.7$\pm$1.8} & \cellcolor{green!30}\textbf{43.3$\pm$0.7} & \cellcolor{yellow!20}16.7$\pm$1.8 & \cellcolor{green!10}{63.3$\pm$0.7} & \cellcolor{green!10}\textbf{28.7$\pm$1.8} \\
& & Pretrain & 57.3 & \textbf{100.0$\pm$0.0} & 78.0$\pm$2.0 & 33.3$\pm$1.8 & \textbf{18.0$\pm$1.2} & \textbf{65.3$\pm$2.7} & 14.0$\pm$0.0 \\
\cmidrule(lr){3-10}
& & Equi Enc & {55.8} & \cellcolor{green!30}\textbf{99.3$\pm$0.7} & \cellcolor{green!30}\textbf{75.3$\pm$2.4} & \cellcolor{green!10}\textbf{32.0$\pm$1.2} & \cellcolor{yellow!20}14.0$\pm$1.2 & \cellcolor{green!10}\textbf{63.3$\pm$1.8} & \cellcolor{green!10}\textbf{26.0$\pm$2.0} \\
& & CNN Enc & 46.7 & 98.0$\pm$0.0 & 72.0$\pm$1.2 & 16.0$\pm$1.2 & \textbf{16.0$\pm$1.2} & 56.7$\pm$1.8 & 4.0$\pm$1.2 \\
\midrule
In-Hand + Ext. & \multirow{2}{*}{Abs Traj} & EquiDiff (Im) & 53.7 & 93.3$\pm$0.7 & 54.7$\pm$5.2 & 25.3$\pm$8.7 & {22.0$\pm$1.2} & 60.0$\pm$2.0 & 15.3$\pm$1.8 \\
Voxel & & EquiDiff (Vo) & 63.9 & 98.7$\pm$0.7 & 74.7$\pm$4.4 & 38.7$\pm$1.3 & 38.7$\pm$0.7 & 64.7$\pm$0.7 & 37.3$\pm$2.7 \\
\midrule
\midrule
Obs & Action & Method & & Hammer Cl. D1 & Mug Cl. D1 & Kitchen D1 & Nut Asse. D0 & Pick Place D0 & Coffee Prep. D1 \\
\midrule
\multirow{4}{*}{\begin{tabular}[l]{@{}l@{}}Large FOV\\In-Hand\end{tabular}} & \multirow{4}{*}{Rel Traj} & Pretrain + FA &  & \cellcolor{green!30}\textbf{76.0$\pm$2.0} & \cellcolor{green!30}{60.0$\pm$2.3} & \cellcolor{green!10}{78.7$\pm$1.8} & \cellcolor{green!30}\textbf{74.7$\pm$1.5} & \cellcolor{green!10}\textbf{50.8$\pm$0.7} & \cellcolor{yellow!20}\textbf{58.0$\pm$2.0} \\
& & Pretrain & & \textbf{76.0$\pm$1.2} & \textbf{62.7$\pm$1.8} & \textbf{80.7$\pm$2.4} & 61.0$\pm$2.1 & 48.2$\pm$1.2 & 50.0$\pm$3.1 \\
\cmidrule(lr){3-10}
& & Equi Enc &  & \cellcolor{green!10}\textbf{67.3$\pm$2.9} & \cellcolor{green!30}\textbf{58.7$\pm$1.8} & \cellcolor{green!10}\textbf{72.0$\pm$3.1} & \cellcolor{green!10}\textbf{69.3$\pm$0.9} & \cellcolor{green!10}\textbf{47.2$\pm$2.8} & \cellcolor{yellow!20}\textbf{44.7$\pm$2.7} \\
& & CNN Enc & &  60.7$\pm$0.7 & 50.7$\pm$1.8 & 68.7$\pm$3.7 & 44.7$\pm$3.3 & 42.7$\pm$1.6 & 30.7$\pm$1.8 \\
\midrule
In-Hand + Ext. & \multirow{2}{*}{Abs Traj} & EquiDiff (Im) & & 65.3$\pm$0.7 & 49.3$\pm$0.7 & 67.3$\pm$0.7 & {74.0$\pm$1.2} & 41.7$\pm$3.2 & {76.7$\pm$0.7} \\
Voxel & & EquiDiff (Vo) &  & 70.0$\pm$2.0 & 52.7$\pm$1.3 & 85.3$\pm$0.7 & 67.3$\pm$0.9 & 57.7$\pm$1.8 & 80.0$\pm$1.2 \\
\bottomrule
\end{tabular}
\end{table*}

\section{Discussion} 
\label{sec:conclusion}

In this paper, we present a practical guide for incorporating symmetry in diffusion policies, achieving performance competitive with or exceeding fully equivariant architectures while requiring significantly less implementation complexity. Notably, our method performs only 2.5\% below voxel-based EquiDiff, despite using only a single eye-in-hand RGB image compared to EquiDiff's four RGBD cameras. Our approach not only defines a new state-of-the-art performance for RGB eye-in-hand diffusion policy, but more importantly, it addresses the trade-off between architectural complexity and sample efficiency when introducing symmetries into policy learning. 

Concretely, we investigate three straightforward approaches for incorporating symmetry: invariant representations through relative trajectory and eye-in-hand perception, integrating equivariant vision encoders, and using Frame Averaging with pretrained encoders. Our extensive experimental evaluation across 12 manipulation tasks in MimicGen yields several important findings. First, we demonstrate that relative trajectory actions consistently outperform absolute trajectory, confirming our theoretical analysis that relative trajectory induces $\SE(3)$-invariance. This finding is particularly valuable because a simple coordinate frame change in action representation can bring a 5-7\% improvement. Second, we found that incorporating symmetry through equivariant vision encoders significantly enhances performance by 9.1\%, highlighting the value of symmetry-aware features while avoiding complex end-to-end reasoning. Lastly, we show that Frame Averaging provides an elegant solution for leveraging the power of pre-trained vision encoders while maintaining equivariance. 

\subsection{Limitations}
There are several limitations of this work that suggest directions for future research. First, only leveraging an eye-in-hand image assumes a good coverage of the entire workspace (thus we use an enlarged FOV in our experiments); however, as shown in Figure~\ref{fig:failure_case}, the limited view still constitutes the most significant failure mode of our system. In future works, this could be addressed by using a fish-eye camera~\cite{chi2024universal}, or a memory mechanism to maintain context across timesteps. Second, while our approaches are theoretically applicable to other policy learning frameworks beyond diffusion models, such as ACT~\cite{aloha}, we limited our investigation to diffusion policies and only experimented in the MimicGen~\cite{mimicgen} and Robomimic~\cite{robomimic} benchmarks. Third, leveraging an equivariant encoder, especially with Frame Averaging, could be computationally expensive. Our method roughly takes twice the GPU hours to train compared with the original Diffusion Policy, but is twice as fast as EquiDiff (Im).
Finally, although our method is well-suited for real-world deployment on systems like UMI~\cite{chi2024universal} , we have not yet demonstrated this transfer to physical robots.

\section*{Acknowledgment}
This work was supported in part by NSF grants 2107256, 2134178, 2314182, 2409351, 2442658, and NASA grant 80NSSC19K1474.



\bibliographystyle{plainnat}
\bibliography{references}

\clearpage

\appendix

\section{Proof of Proposition~\ref{prop:action}}
\label{app:proof_action}
\begin{proof}

{1. Absolute trajectory equivariance:}
Given an absolute trajectory action $a = \{A_{t+i}\}_{i=0}^{n-1}$, each pose $A_{t+i}$ is defined in the world frame. Under the transformation $g$, each pose transforms as $gA_{t+i}, i=0,\dots, n-1$. Thus, by definition, the absolute trajectory transforms equivariantly:
\[
g\cdot a \;=\;\{gA_{t},\,gA_{t+1},\dots\}
\]

{2. Relative trajectory invariance:}
Consider a relative trajectory action $a^r = \{A_{t+i}^{r}\}_{i=0}^{n-1}$ defined in the local gripper frame at the initial time step $t$. The corresponding absolute poses are obtained as $A_{t+i} = T_t A_{t+i}^{r}$. Under the global transform $g$, the absolute pose becomes $g \cdot A_{t+i} = gT_t A_{t+i}^{r}$. Since the relative pose $A_{t+i}^{r}$ appears as a right multiplication factor, it remains unchanged under the global transform. Hence, we have invariance:
\[
g\cdot a^r = a^r
\]
{3. Delta trajectory invariance:}
For a delta trajectory action $a^d = \{A_{t+i}^{d}\}_{i=0}^{n-1}$, each incremental transform $A_{t+i}^{d}$ is expressed in the gripper's local frame at time $t+i-1$. The absolute pose reconstruction is given by $A_{t+i} = T_t \prod_{j=0}^{i-1} A_{t+j}^{d}$. Under the global transform $g$, we have $g\cdot A_{t+i} = gT_t \prod_{j=0}^{i-1} A_{t+j}^{d}$, where each incremental transform $A_{t+j}^{d}$ is multiplied on the right and thus remains unaffected by the global transformation $g$. Therefore, the delta trajectory action is invariant:
\[
g\cdot a^d = a^d
\]
\end{proof}

\section{Proof of Proposition~\ref{prop:policy}}
\label{app:proof_policy}
\begin{proof}
We treat the two cases in parallel.  In both, the policy 
\[
\pi(\ob_t)\;=\;\bigl\{A_{t+i}\bigr\}_{i=0}^{n-1}
\]
outputs a sequence of absolute poses $A_{t+i}\in \SE(3)$.  Internally it first predicts a “local’’
sequence $\bar\pi(\ob_t)=\{A^{r}_{t+i}\}_{i=0}^{n-1}$ or $\bar\pi(\ob_t)=\{A^{d}_{t+i}\}_{i=0}^{n-1}$ (either relative or delta) and then
reconstructs absolute poses by anchoring to the current gripper pose~$T_t$.

{Case 1: Relative trajectories.}
By Definition~\ref{def:rel-traj}, 
\[
A_{t+i}
\;=\;
T_t\,A_{t+i}^{r},
\quad i=0,\dots,n-1,
\]
where $A^r_{t+i}$ is the $i$th pose in the relative sequence
$\bar\pi(\ob_t)=\{A^r_{t+i}\}$.  Thus
\[
\pi(\ob_t)
\;=\;\bigl\{\,T_t\,A^r_{t+i}\bigr\}_{i=0}^{n-1}.
\]
By assumption, $\bar{\pi}$ does not depend on the gripper pose $T_t$ explicitly. Thus, applying the transformation $g$ to the observation has no effect on the relative trajectory prediction:
\begin{equation}
\label{eqn:rel_inv}
\bar{\pi}(g\cdot \ob_t) = \bar{\pi}(I_t, gT_t, w_t) = \bar{\pi}(I_t, T_t, w_t) = \bar{\pi}(\ob_t).
\end{equation}
Reconstructing absolute poses from the
transformed observation gives, for each $i$,
\[
\bigl[\pi(g\cdot\ob_t)\bigr]_{i}
=
(gT_t)\,A^r_{t+i}
=
g\bigl(T_t\,A^r_{t+i}\bigr)
=
g\cdot A_{t+i} = g\cdot\bigl[\pi(\ob_t)\bigr]_{i}
\]
Because this holds for all $i=0,\dots,n-1$, we conclude
\[
\pi\bigl(g\cdot\ob_t\bigr)
=
g\cdot\pi(\ob_t).
\]

{Case 2: Delta trajectories.}
By Definition, the delta-reconstruction is
\[
A_{t+i}
\;=\;
T_t
\;\Bigl(\prod_{j=0}^{i-1}A^d_{t+j}\Bigr),
\quad
i=0,\dots,n-1,
\]
where $\{A^d_{t+j}\}$ is the delta‐sequence from $\bar\pi(\ob_t)$.  Again invariance of $\bar\pi$
under $g$ gives $\bar\pi(g\cdot\ob_t)=\bar\pi(\ob_t)$, so  
\[
\bigl[\pi(g\cdot\ob_t)\bigr]_{i}
=
(gT_t)\,\Bigl(\prod_{j=0}^{i-1}A^d_{t+j}\Bigr)
=
g\bigl[T_t\,(\prod_{j=0}^{i-1}A^d_{t+j})\bigr]
=
g\cdot A_{t+i} = g\cdot \bigl[\pi(\ob_t)\bigr]_{i}
\]
Hence once again 
\[
\pi\bigl(g\cdot\ob_t\bigr)
=
g\cdot\pi(\ob_t).
\]
In both cases we have shown that the entire trajectory satisfies $\pi(g\cdot\ob_t)=g\cdot\pi(\ob_t)$,
i.e.\ $\pi$ is $\SE(3)$‐equivariant.
\end{proof}

\section{Training Detail}
We follow the training setup and hyper-parameters of the prior works~\citep{diffpo,wang2024equivariant,chi2024universal}. Specifically, our RGB observation has a size of $3\times 84\times 84$ (which will be random cropped to $3\times 76\times 76$ during training), and all tasks have a full 6 DoF SE(3) action space. The observation contains two steps of history observation, and the output of the denoising process is a sequence of 16 action steps. We use all 16 steps for training but only execute eight steps in evaluation. In all pretraining encoder variations, we use two steps of proprioceptive observation but only one step of visual observation, following~\citet{chi2024universal}. The vision encoder's output dimension is 64 for for CNN Enc (following~\cite{diffpo}), 128$\times$8 for Equi Enc (128 channel regular representation of $C_8$, following~\cite{wang2024equivariant}), and 512 for Pretrain (following~\cite{chi2024universal}). The diffusion UNet has [512, 1024, 2048] hidden channels for end-to-end training variations (following~\citet{diffpo}), and [256, 512, 1024] hidden channels for pretraining encoder variations (following~\citet{chi2024universal}). We train our models with the AdamW~\citep{adamw} optimizer (with a learning rate of $10^{-4}$ and weight decay of $10^{-6}$) and Exponential Moving Average (EMA). We use a cosine learning rate scheduler with 500 warm-up steps. We use DDPM~\citep{ddpm} with 100 denoising steps for both training and evaluation. We perform training for 600 epochs, and evaluate the method every 10 episodes (60 evaluations in total). All trainings are performed on a single GPU, where we perform training on internal clusters and desktops with different GPU models. Each training of the Pretrain + FA method takes from 3 hours (Stack D1) to 24 hours (Pick Place D0), due to the different sizes of the dataset. The total amount of compute used in this project is roughly 3000 GPU hours.

\section{Robomimic Experiment}
\label{app:robomimic}

\begin{table*}[t]
\setlength\tabcolsep{3pt}
\centering
\scriptsize
\begin{tabular}{lllcccccccc}
\toprule
& & &  & \multicolumn{2}{c}{Lift} & \multicolumn{2}{c}{Can} & \multicolumn{2}{c}{Square} & tool hang \\
\cmidrule(lr){5-6} \cmidrule(lr){7-8} \cmidrule(lr){9-10} \cmidrule(lr){11-11}
Obs & Action & Method & Mean & 100 PH & 100 MH & 100 PH & 100 MH & 100 PH & 100 MH & 100 PH \\\midrule
Large FOV In-Hand & Rel Traj & Ours & \textbf{93.4}  & \textbf{100.0$\pm$0.0} & \textbf{100.0$\pm$0.0} & 99.3$\pm$0.7 & \textbf{100.0$\pm$0.0} & \textbf{88.0$\pm$2.3} & \textbf{92.0$\pm$0.0} & 74.7$\pm$2.4   \\
Voxel & Abs Traj & EquiDiff &  90.4 & \textbf{100.0$\pm$0.0} & \textbf{100.0$\pm$0.0} & 99.3$\pm$0.7 & 96.7$\pm$0.7  & 84.0$\pm$1.2 & 76.7$\pm$1.3 & \textbf{76.0$\pm$0.0}  \\
In-Hand + External & Abs Traj & DiffPo & 87.9 & \textbf{100.0$\pm$0.0} & \textbf{100.0$\pm$0.0} & \textbf{100.0$\pm$0.0} & 95.3$\pm$0.7 & 85.3$\pm$0.7 & 70.7$\pm$0.7 & 64.0$\pm$5.8  \\\bottomrule
\end{tabular}
\caption{The performance of our method compared with the baselines in Robomimic. We experiment with 100 Proficient-Human (PH) or Multi-Human (MH) demos in each environment. Results averaged over three seeds. $\pm$ indicates standard error.}
\label{tab:sim_robomimic}
\end{table*}

In this section, we perform an experiment in the Robomimic~\cite{robomimic} environments. We compare our Pretrain + FA method against EquiDiff~\cite{wang2024equivariant} and the vanilla Diffusion Policy~\cite{diffpo}. As shown in Table~\ref{tab:sim_robomimic}, our method generally outperforms the baselines, achieving an average improvement of 3\% compared with EquiDiff.

\begin{table*}[t]
\scriptsize
\setlength\tabcolsep{1.5pt}
\centering
\setlength{\fboxsep}{1pt}
\caption{The performance comparison of pretrained encoder with Frame Averaging, pretrained encoder, and no pretraining. All policies are trained using 100 expert demonstrations. We perform 60 evaluations (each with 50 policy rollouts) throughout the training, and report the best average success rate. Results averaged over three seeds, $\pm$ indicates standard error. \colorbox{green!30}{Dark green box} indicates outperforming EquiDiff with voxel inputs; \colorbox{green!10}{Light green box} indicates outperforming EquiDiff with image inputs; \colorbox{yellow!20}{Yellow box} indicates underperforming both. \textbf{Bold} indicates whether the symmetric encoder outperforms the non-symmetric version in the corresponding training setting. Performance of EquiDiff is reported by~\cite{wang2024equivariant}.}
\label{tab:equi_enc_ext}
\begin{tabular}{l l l c c c c c c c}
\toprule
Obs & Action & Method & Mean & Stack D1 & Stack Three D1 & Square D2 & Threading D2 & Coffee D2 & Three Pc. D2 \\
\midrule
\multirow{3}{*}{In-Hand + Ext.} & \multirow{3}{*}{Rel Traj} & Pretrain + FA & 64.9 & \cellcolor{green!30}\textbf{100.0$\pm$0.0} & \cellcolor{green!30}\textbf{80.7$\pm$0.7} & \cellcolor{green!30}\textbf{42.7$\pm$1.3} & \cellcolor{green!10}\textbf{28.0$\pm$1.2} & \cellcolor{green!30}\textbf{72.7$\pm$3.3} & \cellcolor{green!10}\textbf{32.0$\pm$2.3} \\
& & Pretrain & 58.2 & \textbf{100.0$\pm$0.0} & 72.7$\pm$2.4 & 33.3$\pm$1.3 & 22.0$\pm$2.0 & 63.3$\pm$0.7 & 18.0$\pm$2.0 \\
& & No Pretrain & 49.4 & 89.3$\pm$2.7 & 52.7$\pm$3.7 & {20.7$\pm$1.3} & {18.7$\pm$3.5} & {63.3$\pm$1.3} & 3.3$\pm$0.7 \\
\midrule
\multirow{3}{*}{\begin{tabular}[l]{@{}l@{}}Large FOV\\In-Hand\end{tabular}} & \multirow{3}{*}{Rel Traj} & Pretrain + FA & {61.4} & \cellcolor{green!30}\textbf{100.0$\pm$0.0} & \cellcolor{green!30}\textbf{86.7$\pm$1.8} & \cellcolor{green!30}\textbf{43.3$\pm$0.7} & \cellcolor{yellow!20}16.7$\pm$1.8 & \cellcolor{green!10}{63.3$\pm$0.7} & \cellcolor{green!10}\textbf{28.7$\pm$1.8} \\
& & Pretrain & 57.3 & \textbf{100.0$\pm$0.0} & 78.0$\pm$2.0 & 33.3$\pm$1.8 & \textbf{18.0$\pm$1.2} & \textbf{65.3$\pm$2.7} & 14.0$\pm$0.0 \\
& & No Pretrain & 46.7 & 98.0$\pm$0.0 & 72.0$\pm$1.2 & 16.0$\pm$1.2 & {16.0$\pm$1.2} & 56.7$\pm$1.8 & 4.0$\pm$1.2 \\
\midrule
In-Hand + Ext. & \multirow{2}{*}{Abs Traj} & EquiDiff (Im) & 53.7 & 93.3$\pm$0.7 & 54.7$\pm$5.2 & 25.3$\pm$8.7 & {22.0$\pm$1.2} & 60.0$\pm$2.0 & 15.3$\pm$1.8 \\
Voxel & & EquiDiff (Vo) & 63.9 & 98.7$\pm$0.7 & 74.7$\pm$4.4 & 38.7$\pm$1.3 & 38.7$\pm$0.7 & 64.7$\pm$0.7 & 37.3$\pm$2.7 \\
\midrule
\midrule
Obs & Action & Method & & Hammer Cl. D1 & Mug Cl. D1 & Kitchen D1 & Nut Asse. D0 & Pick Place D0 & Coffee Prep. D1 \\
\midrule
\multirow{3}{*}{In-Hand + Ext.} & \multirow{3}{*}{Rel Traj} & Pretrain + FA &  & \cellcolor{green!10}\textbf{67.3$\pm$1.8} & \cellcolor{green!30}\textbf{55.3$\pm$1.3} & \cellcolor{green!10}\textbf{82.7$\pm$0.7} & \cellcolor{green!30}\textbf{78.0$\pm$1.5} & \cellcolor{green!30}\textbf{61.8$\pm$0.6} & \cellcolor{green!10}78.0$\pm$1.2 \\
& & Pretrain &  & 61.3$\pm$0.7 & 54.0$\pm$1.2 & 76.0$\pm$1.2 & 72.3$\pm$1.7 & 55.5$\pm$3.8 & 69.3$\pm$1.3 \\
& & No Pretrain &  & 58.7$\pm$0.7 & 46.7$\pm$0.7 & 62.0$\pm$2.0 & {57.3$\pm$0.9} & 39.8$\pm$1.2 & \textbf{80.0$\pm$2.0} \\
\midrule
\multirow{3}{*}{\begin{tabular}[l]{@{}l@{}}Large FOV\\In-Hand\end{tabular}} & \multirow{3}{*}{Rel Traj} & Pretrain + FA &  & \cellcolor{green!30}\textbf{76.0$\pm$2.0} & \cellcolor{green!30}{60.0$\pm$2.3} & \cellcolor{green!10}{78.7$\pm$1.8} & \cellcolor{green!30}\textbf{74.7$\pm$1.5} & \cellcolor{green!10}\textbf{50.8$\pm$0.7} & \cellcolor{yellow!20}\textbf{58.0$\pm$2.0} \\
& & Pretrain & & \textbf{76.0$\pm$1.2} & \textbf{62.7$\pm$1.8} & \textbf{80.7$\pm$2.4} & 61.0$\pm$2.1 & 48.2$\pm$1.2 & 50.0$\pm$3.1 \\
& & No Pretrain & &  60.7$\pm$0.7 & 50.7$\pm$1.8 & 68.7$\pm$3.7 & 44.7$\pm$3.3 & 42.7$\pm$1.6 & 30.7$\pm$1.8 \\
\midrule
In-Hand + Ext. & \multirow{2}{*}{Abs Traj} & EquiDiff (Im) & & 65.3$\pm$0.7 & 49.3$\pm$0.7 & 67.3$\pm$0.7 & {74.0$\pm$1.2} & 41.7$\pm$3.2 & {76.7$\pm$0.7} \\
Voxel & & EquiDiff (Vo) &  & 70.0$\pm$2.0 & 52.7$\pm$1.3 & 85.3$\pm$0.7 & 67.3$\pm$0.9 & 57.7$\pm$1.8 & 80.0$\pm$1.2 \\
\bottomrule
\end{tabular}
\end{table*}

\section{Pretraining and Frame Averaging with External View}
In this experiment, we extend our analysis from Section~\ref{sec:exp_equi_enc} to the external view setting to verify whether our findings generalize across different observation configurations. Specifically, we perform an additional experiment on using pretrained encoders (with and without Frame Averaging) with In-Hand + External view. Similar to Section~\ref{sec:exp_equi_enc}, we consider three methods in each view setting: 1) \emph{No Pretrain}: A standard ResNet-18~\cite{resnet} trained from scratch; 2) \emph{Pretrain}: A standard ResNet-18 pretrained on ImageNet-1k~\cite{deng2009imagenet}, without explicit symmetry constraints; 3) \emph{Pretrain + FA}: A pretrained ResNet-18 enhanced with Frame Averaging (Equation~\ref{eqn:frame_avg}) to achieve $C_8$-equivariance without modifying the underlying network architecture.

As shown in Table~\ref{tab:equi_enc_ext},  the benefits of Frame Averaging remain consistent across both observation settings. In the In-Hand + External view setting, Pretrain + FA yields a 6.7\% improvement compared with not using Frame Averaging (Pretrain), and a 15.5\% improvement compared with training from scratch (No Pretrain). Notably, Pretrain + FA in In-Hand + External view outperforms EquiDiff (Im) in all tasks, and even achieves a 1\% higher average performance compared with EquiDiff (Vo). This is particularly impressive considering the additional complexity of EquiDiff, as discussed in Section~\ref{sec:exp_equi_enc}. Comparing Pretrain + FA across different views, despite using an additional camera, In-Hand + External only outperforms Large FOV In-Hand by 2.5\%. This finding verifies our analysis in Section~\ref{sec:inv_policy}, and suggests Large FOV In-Hand is preferable in many applications due to its simplicity (requiring only a single eye-in-hand camera) and easier transferability to diverse robotic platforms beyond tabletop manipulation scenarios.


\section{Pretraining and Frame Averaging with Absolute Action and Multi-Camera Observation}

To demonstrate generality beyond the in-hand setup, we evaluate the proposed Pretrain+FA encoder under the same observation/action setting as Diffusion Policy and EquiDiff (in-hand + external views and absolute actions) in four different environments. As shown in Table~\ref{tab:fa_abs}, employing the Pretrain+FA encoder yields a significant 21.5\% and 9.7\% improvement for Diffusion Policy and EquiDiff, respectively. These results confirm the encoder's plug-and-play nature across observation and action parameterizations.

\begin{table*}[t]
\scriptsize
\setlength\tabcolsep{5pt}
\centering
\setlength{\fboxsep}{1pt}
\caption{Validating the Pretrain+FA encoder in absolute trajectory and multi-camera observations.}
\label{tab:fa_abs}
\begin{tabular}{l l l c c c c c}
\toprule
Obs & Action & Method & Mean & Stack Three D1 & Square D2 & Coffee D2 & Nut Assembly D0 \\
\midrule
\multirow{4}{*}{In-Hand + Ext.} & \multirow{4}{*}{Abs Traj} & DP & 36.2 & 38.0$\pm$0.0 & 8.0$\pm$1.2 & 44.0$\pm$1.2 & 54.7$\pm$2.3 \\
 &  & DP + Pretrain + FA & 57.7 & 58.7$\pm$1.2 & 38.0$\pm$9.2 & 56.0$\pm$2.0 & 78.0$\pm$2.6 \\
\cmidrule(lr){3-8}
&  & EquiDiff & 53.5 & 54.7$\pm$5.2 & 25.3$\pm$8.7 & 60.0$\pm$2.0 & 74.0$\pm$1.2 \\
&  & EquiDiff + Pretrain + FA & 63.2 & 68.7$\pm$2.3 & 29.3$\pm$3.0 & 78.0$\pm$2.0 & 76.7$\pm$3.2 \\
\bottomrule
\end{tabular}
\end{table*}

\section{Ablation Study}

\subsection{Isolating Relative Trajectory and Symmetric Feature Extraction}

\begin{table*}[ht]
\scriptsize
\setlength\tabcolsep{5pt}
\centering
\setlength{\fboxsep}{1pt}
\caption{Ablation isolating the contributions of (i) relative trajectory and (ii) symmetric feature extraction (Pretrain+FA).}
\label{tab:ablation}
\begin{tabular}{l c c c c c c c}
\toprule
Obs & Rel Traj & Pretrain + FA & Mean & Stack Three D1 & Square D2 & Coffee D2 & Nut Asse. D0 \\
\midrule
\multirow{3}{*}{\begin{tabular}[l]{@{}l@{}}Large FOV\\In-Hand\end{tabular}}  & \textcolor{ggreen}{\Checkmark} & \textcolor{ggreen}{\Checkmark} & 67.0 & 86.7$\pm$1.8 & 43.3$\pm$0.7 & 63.3$\pm$0.7  & 74.7$\pm$1.5 \\
 & \textcolor{gred}{\XSolidBrush} & \textcolor{ggreen}{\Checkmark} & 55.9 & 68.7$\pm$2.3 & 26.7$\pm$11.5 & 60.0$\pm$0.0 & 68.0$\pm$3.0 \\
 & \textcolor{ggreen}{\Checkmark} & \textcolor{gred}{\XSolidBrush} & 47.4 & 72.0$\pm$1.2 & 16.0$\pm$1.2 & 56.7$\pm$1.8  &  44.7$\pm$3.3\\
\bottomrule
\end{tabular}
\end{table*}

We explicitly isolate each component in our design. As shown in Table~\ref{tab:ablation}, starting from the full model, replacing the relative trajectory with absolute trajectory reduces performance by 11.1\% across four tasks, while replacing the encoder with a non-pretrained CNN reduces performance more significantly by 19.6\%. This result highlights the complementary benefits from symmetry-aware features and invariant action parameterization. 

\subsection{Ablating Proprioception}

We test Proposition~\ref{prop:policy}'s assumption by removing the gripper pose from the policy input in Table~\ref{tab:ablate-proprio}. As expected, this tighter symmetry assumption reduces performance—indicating that proprioception, while symmetry-breaking, provides valuable context. 

\begin{table*}[ht]
\scriptsize
\centering
\caption{Ablating proprioception input in the policy}
\label{tab:ablate-proprio}
\begin{tabular}{lcc}
\toprule
 & {Stack Three D1} & {Square D2} \\
\midrule
With proprioception & 72.0 & 16.0 \\
No proprioception & 64.7 & 14.7 \\
\bottomrule
\end{tabular}
\end{table*}

\subsection{Ablating Symmetry Group}

We vary the SO(2) discretization used by the equivariant encoder. As shown in Table~\ref{tab:ablate-group}, reducing the cyclic group order degrades performance (C8 > C4 > C2), aligning with prior observations~\cite{e2cnn,corl22} that C8 is a practical sweet spot for 2D rotational symmetry.

\begin{table*}[ht]
\scriptsize
\centering
\caption{Effect of SO(2) discretization.}
\label{tab:ablate-group}
\begin{tabular}{lcc}
\toprule
{Group} & {Stack Three D1} & {Square D2} \\
\midrule
C8 & 75.3 & 32.0 \\
C4 & 68.0 & 30.7 \\
C2 & 63.3 & 20.7 \\
\bottomrule
\end{tabular}
\end{table*}

\section{EquiDiff with Large FOV In-Hand Only}

In this experiment, we evaluate EquiDiff with only a large-FOV eye-in-hand camera setting. The comparison is shown in Table~\ref{tab:equi_ih}. Compared to the original external-camera configuration, EquiDiff's mean success drops by 12.8\%, underscoring the benefit EquiDiff derives from external viewpoint signals. This result also justifies our choice of using the original EquiDiff observation setting in our main results.

\begin{table*}[h!]
\scriptsize
\setlength\tabcolsep{0.9pt}
\centering
\setlength{\fboxsep}{1pt}
\caption{Performance of EquiDiff under a large FOV in-hand-only camera setting vs. the original in-hand + external and voxel (multi-RGBD) settings. Removing external cameras substantially reduces EquiDiff performance on several tasks.}
\label{tab:equi_ih}
\begin{tabular}{l l l c c c c c c c}
\toprule
Obs & Action & Method & Mean & Stack D1 & Stack Three D1 & Square D2 & Threading D2 & Coffee D2 & Three Pc. D2 \\
\midrule
Large FOV In-Hand & \multirow{3}{*}{Abs Traj} & EquiDiff (Im) & 40.9 & 96.0$\pm$0.0 & 61.3$\pm$5.0 & 8.7$\pm$1.2 & 13.3$\pm$2.3 & 47.3$\pm$3.1 & 3.3$\pm$1.2 \\
In-Hand + Ext. &   & EquiDiff (Im) & 53.7 & 93.3$\pm$0.7 & 54.7$\pm$5.2 & 25.3$\pm$8.7 & {22.0$\pm$1.2} & 60.0$\pm$2.0 & 15.3$\pm$1.8 \\
Voxel & & EquiDiff (Vo) & 63.9 & 98.7$\pm$0.7 & 74.7$\pm$4.4 & 38.7$\pm$1.3 & 38.7$\pm$0.7 & 64.7$\pm$0.7 & 37.3$\pm$2.7 \\
\midrule
\midrule
Obs & Action & Method & & Hammer Cl. D1 & Mug Cl. D1 & Kitchen D1 & Nut Asse. D0 & Pick Place D0 & Coffee Pre. D1 \\
\midrule
Large FOV In-Hand & \multirow{3}{*}{Abs Traj} & EquiDiff (Im) &  & 59.3$\pm$4.2 & 50.7$\pm$2.3 & 55.3$\pm$1.2 & 40.0$\pm$4.0 & 27.7$\pm$2.9 & 27.3$\pm$1.2 \\
In-Hand + Ext. &  & EquiDiff (Im) & & 65.3$\pm$0.7 & 49.3$\pm$0.7 & 67.3$\pm$0.7 & {74.0$\pm$1.2} & 41.7$\pm$3.2 & {76.7$\pm$0.7} \\
Voxel & & EquiDiff (Vo) &  & 70.0$\pm$2.0 & 52.7$\pm$1.3 & 85.3$\pm$0.7 & 67.3$\pm$0.9 & 57.7$\pm$1.8 & 80.0$\pm$1.2 \\
\bottomrule
\end{tabular}
\end{table*}

\end{document}